\PassOptionsToPackage{table}{xcolor}
\PassOptionsToPackage{xcdraw}{xcolor}
\documentclass[10pt,twocolumn,letterpaper]{article}

\usepackage{cvpr}              
\usepackage{tabularray}
\usepackage{float}
\usepackage{placeins}
\usepackage{graphicx}
\usepackage{multirow}
\usepackage[table,xcdraw]{xcolor}
\usepackage{bbding}
\definecolor{orange}{HTML}{F9CB9C}
\definecolor{red}{HTML}{EA9999}
\definecolor{yellow}{HTML}{FFFC9E}
\newcommand{\shadeText}[2]{\colorbox{#1}{#2}}
\definecolor{cvprblue}{rgb}{0.21,0.49,0.74}
\usepackage[pagebackref,breaklinks,colorlinks,allcolors=cvprblue]{hyperref}

\def\paperID{12101} 
\def\confName{CVPR}
\def\confYear{2025}

\title{BARD-GS: Blur-Aware Reconstruction of Dynamic Scenes via Gaussian Splatting}

\author{
    Yiren Lu, 
    Yunlai Zhou, 
    Disheng Liu, 
    Tuo Liang, 
    Yu Yin\textsuperscript{\Envelope} \\
    Case Western Reserve University \\
    \{yiren.lu, yunlai.zhou, disheng.liu, tuo.liang, yu.yin\}@case.edu \\
    \url{https://vulab-ai.github.io/BARD-GS/}
}

\begin{document}
\maketitle
\begin{abstract}
3D Gaussian Splatting (3DGS) has shown remarkable potential for static scene reconstruction, and recent advancements have extended its application to dynamic scenes. 
However, the quality of reconstructions depends heavily on high-quality input images and precise camera poses, which are not that trivial to fulfill in real-world scenarios.
Capturing dynamic scenes with handheld monocular cameras, for instance, typically involves simultaneous movement of both the camera and objects within a single exposure. This combined motion frequently results in image blur that existing methods cannot adequately handle.
To address these challenges, we introduce BARD-GS, a novel approach for robust dynamic scene reconstruction that effectively handles blurry inputs and imprecise camera poses. Our method comprises two main components: 1) camera motion deblurring and 2) object motion deblurring. 
By explicitly decomposing motion blur into camera motion blur and object motion blur and modeling them separately, we achieve significantly improved rendering results in dynamic regions.
In addition, we collect a real-world motion blur dataset of dynamic scenes to evaluate our approach. Extensive experiments demonstrate that BARD-GS effectively reconstructs high-quality dynamic scenes under realistic conditions, significantly outperforming existing methods.
\end{abstract}    
\section{Introduction}
\label{sec:intro}
\begin{figure}
    \centering
    \includegraphics[width=1\linewidth]{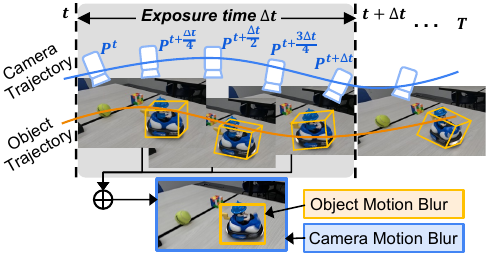}
    \caption{\textbf{The formation process of motion blur.} It originates from two sources: camera-induced blur caused by camera movements during exposure, and object-induced blur resulting from fast objects moving. The static regions of a scene are affected solely by camera motion blur, while dynamic regions are impacted by both camera and object motion blur.}
    \vspace{-5mm}
    \label{fig:motion_blur_formation}
\end{figure}
The emergence of Neural Radiance Fields (NeRF) \cite{mildenhall2020nerf} and 3D Gaussian Splatting (3DGS) \cite{kerbl3Dgaussians} has significantly advanced 3D scene reconstruction, with widespread applications in 3D editing \cite{haque2023instruct, wang2022clip, chen2024gaussianeditor, wang2024gaussianeditor, lu2024view, yin2023nerfinvertor} and SLAM \cite{Matsuki:Murai:etal:CVPR2024, rosinol2023nerf, Sucar:etal:ICCV2021, Zhu2022CVPR, keetha2024splatam}. Recent research has extended NeRF and 3DGS to dynamic scenes \cite{bae2024ed3dgs, yang2024deformable, yang2023gs4d, lu20243d, zhu2024motiongs, duan20244d}, demonstrating promising results. However, their performance declines when processing motion-blurred images.

Motion blur is a common occurrence in everyday photography, typically arising from movements during the exposure time, as demonstrated in Fig.~\ref{fig:motion_blur_formation}. It can be decomposed as camera- and object-induced blur. Beyond the loss of visual clarity, motion blur introduces geometric inconsistencies across frames, posing a significant challenge to existing 3D reconstruction methods.
While prior works have explored motion-blurred reconstruction \cite{wang2023badnerf, zhao2024badgaussians, Chen_deblurgs2024, lee2023exblurf, li2022deblurnerf}, they primarily focus on static scenes, lacking the ability to handle dynamic object motion. Moreover, integrating existing deblur-GS \cite{Chen_deblurgs2024, zhao2024badgaussians} with dynamic-GS frameworks \cite{yang2024deformable, yang2023gs4d} addresses only camera-induced blur, leaving object motion blur unresolved.

There are two NeRF-based approaches \cite{bui2023dyblurf, sun2024dyblurf} trying to eliminate both types of motion blur by optimizing multiple latent rays across exposure time and warping them at different timestamps for dynamic rendering. 
However, these methods couple object and camera motion effects within the latent ray representation. 
Since static region typically dominates the scene, rays tend to adjust primarily for camera motion blur, resulting in poor reconstruction in dynamic areas, especially with severe blur caused by fast-moving objects. 
Besides, the implicit neural representation of NeRF makes it challenging to decouple these effects. 



In this paper, we present \textbf{BARD-GS}, a \underline{\textbf{B}}lur-\underline{\textbf{A}}ware \underline{\textbf{R}}econstruction framework for \underline{\textbf{D}}ynamic scenes based on \underline{\textbf{G}}aussian \underline{\textbf{S}}platting.
To tackle the challenges posed by blurry input, 
we decompose motion blur into two types: \textit{camera motion blur}, modeled by variations in camera pose, and \textit{object motion blur}, modeled by the movement of dynamic Gaussians. By modeling these two types of blur separately, we address the problem in two sequential stages: 1) camera motion deblurring and 2) object motion deblurring.

In the first stage, we address camera motion blur by modeling the camera’s trajectory within each exposure time as a sequence of virtual camera poses. We simulate camera motion blur by rendering images from these virtual poses and combining them to reconstruct blurry images. By aligning the static regions of reconstructed blurry images to the input images, we optimize the 3D Gaussians to sharply reconstruct the static regions of the scene, effectively isolating and eliminating camera-induced blur. Note that this stage focuses exclusively on static regions to better distinguish camera and object motion blur.

In the second stage, we address object motion blur and accurately reconstruct dynamic regions while further refining static areas. We learn a time-conditioned deformation field to track the spatial shifts of 3D Gaussians over time, capturing the motion of dynamic objects. Similar to camera movement modeling, we model the object positions at different timestamps within the exposure time, using the trajectory of 3D Gaussians.
We then render images of dynamic objects at virtual timestamps and combine rendered images to reconstruct object-induced blur. By aligning the entire reconstructed blurry images with the input, we achieve sharp reconstructions for both dynamic and static areas. This explicit separation and modeling of camera- and object-induced blur greatly enhance the rendering quality in dynamic regions compared to previous approaches.



In the absence of existing dynamic scene datasets with motion blur, we collect a real-world dataset to address this gap. Different from synthetic datasets used in previous works, where blurry images are generated by averaging consecutive frames, our dataset captures motion blur that closely aligns with real-world scenarios. We provide paired blurry and sharp images captured from diverse environments; details about the dataset acquisition can be found in Sec.~\ref{subsec:synthesized_data}.
In summary, our key contributions include:
\begin{itemize}
    \item We propose BARD-GS, a blur-aware dynamic scene reconstruction framework, to tackle challenges posed by blurry inputs and imprecise camera poses.
    \item We decouple object motion blur from camera motion blur and model each separately in an explicit manner, resulting in a substantial improvement in reconstruction quality within dynamic regions compared to previous methods.
    \item A new real-world motion blur dataset is proposed for evaluating the novel view synthesis performance under real-world blurry scenarios.
\end{itemize}

\section{Related Work}
\label{sec/2_related_work}
\subsection{Image Deblurring}
The primary objective of image deblurring is to recover a clear image from a blurry one. 
Recently, due to the power of deep learning, image deblurring tasks have mainly focused on learning-based methods to solve the problem.
In single-image deblurring, existing methods primarily achieve deblurring by leveraging CNN-based models \cite{zhang2019deep}, Transformer-based models \cite{zamir2022restormer} or by reconstructing the image with the help of prior knowledge from generative models \cite{kupyn2019deblurganv2}. With the rapid development of deep learning, these works \cite{zhang2019deep,kupyn2018deblurgan,kupyn2019deblurganv2,chen2024hierarchical,zamir2022restormer,tsai2022stripformer} have achieved superior result on single image deblurring.
Although these methods achieve impressive results, they primarily work within the 2D image space and are unable to capture scene geometry in 3D space, leading to limitations in maintaining spatial consistency. 
Our method achieves deblurring in dynamic 3D scenes by physically modeling the blur formation process, demonstrating a better result compared to 3D reconstruction with pre-deblurred images.


\subsection{Dynamic 3D Gaussian Splatting}
3D Gaussian Splatting (3DGS) \cite{kerbl3Dgaussians} has been a popular 3D scene representation, known for its explicit structure and real-time novel view rendering. As originally designed for static scenes, 3DGS lacks capability for dynamic scene modeling. Recently, various researches \cite{bae2024ed3dgs, yang2024deformable, yang2023gs4d, lu20243d, zhu2024motiongs, duan20244d, katsumata2024compact, kratimenos2024dynmf, li2024spacetime, wang2024shape} have attempted to adapt it for dynamic scene reconstruction.
In order to model the dynamic scene, Dynamic 3D Gaussians \cite{luiten2023dynamic} utilize a table to store the information of each 3D Gaussians at every timestamp, including mean and variance of Gaussians. 4D Gaussian Splatting \cite{yang2023gs4d} achieves dynamic scene modeling by directly lifting the original 3D Gaussian into 4D space. Deformable-3D-Gaussians \cite{yang2024deformable} utilize a multi-layer perceptron (MLP) to learn the position, rotation, and scale for each 3D Gaussians at every timestamp. These methods have achieved excellent dynamic scene reconstruction results, however, all of them require high-quality sharp images and precise camera poses as input. Our approach considers the physical process of motion blur formation, enabling it to handle inputs with both camera-induced and object-induced motion blur, resulting in substantially improved reconstruction quality.

\begin{figure*}
    \centering
    \includegraphics[trim=0in 0.1in 0in 0in,clip,width=\linewidth]{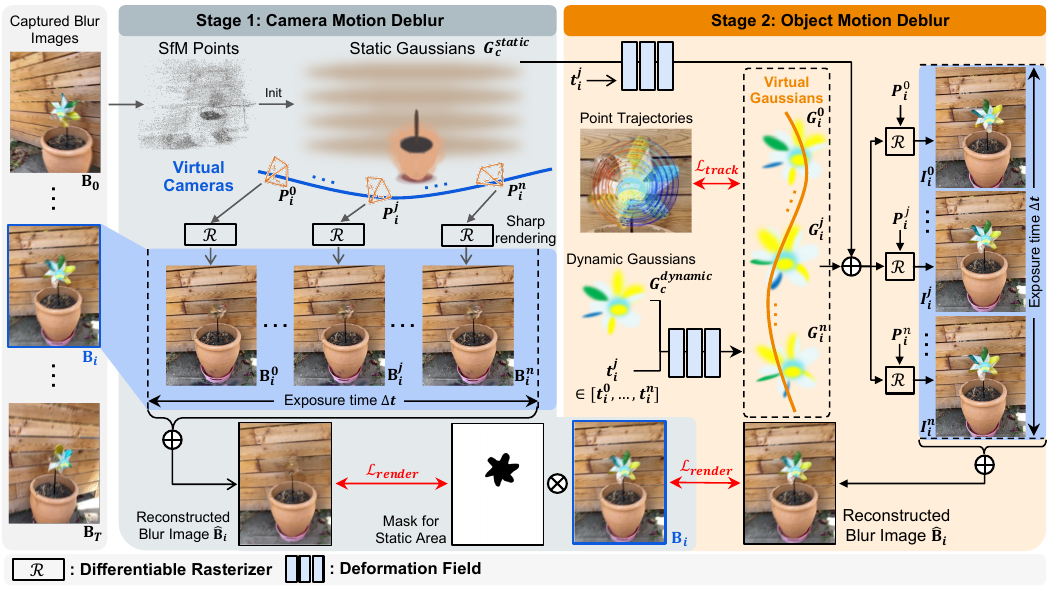}
    \caption{\textbf{An overview of the pipeline.} Our method consists of two stages: camera motion deblur and object motion deblur. In the first stage, we handle camera motion blur by modeling the camera’s trajectory during each exposure, resulting in sharp reconstruction in the static regions. Then, we utilized the optimized camera poses together with the depth map obtain from DepthAnything to initialize the dynamic Gaussians. In the second stage, we address object motion blur by modeling the trajectory of 3D Gaussians within each exposure using deformation field, which allow us to achieve clear reconstruction in the dynamic regions.}
    \label{fig:pipeline}
\end{figure*}

\subsection{3D Reconstruction from Blurry Input}
Recently, there have been many NeRF-based \cite{Lee_2023_CVPR, li2022deblurnerf, wang2023badnerf, peng2022pdrf} and 3DGS-based \cite{zhao2024badgaussians, Chen_deblurgs2024, lee2024smurf, lee2024deblurring, peng2024bags, seiskari2024gaussian, darmon2024robust} methods trying to achieve high-quality 3D reconstruction from blurry input. Deblur-NeRF \cite{li2022deblurnerf} utilizes a Deformable Sparse Kernel (DSK) to model the formation of blur by deforming the canonical kernel. BAD-NeRF \cite{wang2023badnerf} , BAD-Gaussians \cite{zhao2024badgaussians} and Deblur-GS \cite{Chen_deblurgs2024} model the camera trajectory during the exposure time, generating multiple sharp images along the path and blending them to simulate the process of blur formation. These methods can produce high-quality rendering results for static scenes, however, none of them is designed to handle dynamic scenes. 
There are two works share the same goal with us \cite{bui2023dyblurf, sun2024dyblurf} and both of them are NeRF-based. They share a same idea to generate multiple latent rays within the exposure time to model the blur formation process. The rays are then twisted to perform volume rendering on different points across various timestamps, allowing for dynamic modeling.
In this approach, the blurring effects from both camera and object motion are combined within a single ray warping process, causing them to be interdependent. As a result, the static regions dominate, leading to poor detail representation in dynamic areas. Our method explicitly decouples these two sources of blur by modeling the motion of camera and the motion of dynamic Gaussians. By handling them sequentially, our method achieves significantly improved rendering results in dynamic regions.


\section{Method}
\label{sec:method}
We introduce BARD-GS, a blur-aware dynamic scene reconstruction method based on Gaussian Splatting, designed to reconstruct scenes from blurry input. 
Our approach explicitly separates motion blur into two components: \textit{camera motion blur} and \textit{object motion blur}. These are modeled independently using the trajectories of cameras and 3D Gaussians, respectively.
In Sec.~\ref{subsec:preliminary} and~\ref{subsec:motion_blur_formation}, we provide an overview of 3DGS and the formation of motion blur, respectively. We then describe the two main components of our pipeline: \textbf{1)} camera motion deblurring (Sec.~\ref{subsec:static_deblur}) and \textbf{2)} object motion deblurring (Sec.~\ref{subsec:dynamic_deblur}), illustrating how each type of blur is effectively addressed.

\subsection{Preliminary: 3D Gaussian Splatting}
\label{subsec:preliminary}
3D Gaussian Splatting (3DGS) \cite{kerbl3Dgaussians} represents a scene using 3D Gaussian ellipsoids, each parameterized by a center $\mathbf{x}$, covariance matrix $\boldsymbol{\Sigma}$, opacity $\alpha$ and learnable spherical harmonics for color.
During differentiable rasterization, 3D Gaussians are first projected into 2D image space with a given camera pose.
Then each pixel color $\hat{c}(\mathbf{x})$ is calculated based on the overlapping Gaussians projected onto it, using the alpha-blending process:

\begin{equation}
    \hat{c}(\mathbf{x})=\sum_{i \in \mathcal{N}} c_i \alpha_i(\mathbf{x} )\prod_{j=1}^{i-1}\left(1-\alpha_j(\mathbf{x})\right),
\end{equation}
where $c_i$ is the color of each projected Gaussian given by spherical harmonics and $\alpha_i$ is the opacity multiplies with the projected covariance matrix $\boldsymbol{\Sigma}^{\prime}$.
During reconstruction, the parameters of 3D Gaussians are optimized by minimizing the pixel-wise L1 loss together with SSIM loss between the rendered image $\hat{I}$ and the ground truth image $I_{gt}$.

\subsection{Motion Blur Formation}
\label{subsec:motion_blur_formation}
Motion blur results from the movement of cameras or object within an exposure time. In the perspective of physics, due to the above-mentioned movement, the photons from a single object will accumulate across different locations on the sensor, resulting in a ``trail" or streaked path that corresponds to the motion trajectory. The physical process can be described by the following equation:

\begin{equation}
   \mathbf{B}(\mathbf{u}, \mathbf{v})=\phi \int_0^\tau \mathbf{S}_{\mathrm{t}}(\mathbf{u}, \mathbf{v}) \mathrm{dt}, 
\end{equation}
where $\mathbf{S}_{\mathrm{t}}(\mathbf{u}, \mathbf{v})$ is the instantaneous radiance of the scene at pixel location $(\mathbf{u}, \mathbf{v})$ at a specific time $\mathrm{t}$ within an exposure time $\tau$. $\mathbf{B}(\mathbf{u}, \mathbf{v})$ is the resulting blurred pixel at $(\mathbf{u}, \mathbf{v})$, calculated by integrating instantaneous radiance over an exposure duration, ``averaging" the positions of objects in motion, and $\phi$ is a normalization factor. To facilitate the simulation of this process, we utilize a discretized version of the above equation:

\begin{equation}
\mathbf{B}(\mathbf{u}, \mathbf{v}) \approx \frac{1}{n} \sum_{i=0}^{n-1} \mathbf{S}_i(\mathbf{u}, \mathbf{v}).
\end{equation}
Here, the blurry image $\mathbf{B}$ is approximated by averaging $n$ discrete sharp images $\mathbf{S}_i$ captured within an exposure time.

\subsection{Camera Motion Modeling and Deblurring}
In camera motion deblurring, we will handle motion blur caused by camera movement and reconstruct static part of the scene. Inspired by \cite{zhao2024badgaussians}, we compute a set of corresponding virtual camera poses for each real camera to represent the camera movement within an exposure time. 
Here, we refer to the camera associated with each input image as a \textit{real camera}, and the cameras interpolated within the exposure time as \textit{virtual cameras}.
These virtual cameras are used to render virtual sharp images, which will be averaged to reconstruct the blurry input. 

\label{subsec:static_deblur}
\subsubsection{Camera Optimization and Trajectory Modeling}
\label{subsubsec:camera_optimization}
Given a set of $T$ blurry images $\{\mathbf{B}_i\}_{i=1}^T$ as input, we utilize off-the-shelf SfM method COLMAP \cite{schoenberger2016sfm, schoenberger2016mvs} to get initial inaccurate camera poses $\{\widetilde{\mathbf{P}}_i\}_{i=1}^T$. 
For each real camera pose $\widetilde{\mathbf{P}}_i$, the corresponding set of virtual cameras is denoted as $\{\mathbf{P}_i^j\}_{j=0}^{n}$, $\mathbf{P}_i^j$ represents the $j$th virtual camera associate with the $i$th blurry image. Each $\mathbf{P}_i^j$ consists of $\mathbf{R}_i^j$ and $\mathbf{t}_i^j$, representing rotation and translation. These virtual cameras are interpolated from $\mathbf{P}_i^0$ and $\mathbf{P}_i^{n}$, which represent the poses at the beginning and the end of the exposure period. $\mathbf{P}_i^0$ and $\mathbf{P}_i^{n}$ are calculated by applying a learnable delta value to the inaccurate pose $\widetilde{\mathbf{P}}_i$.
\begin{equation}
    \mathbf{P}_i^0 = \widetilde{\mathbf{P}}_i \cdot \Delta \mathbf{P}_i^0, \quad \mathbf{P}_i^n = \widetilde{\mathbf{P}}_i \cdot \Delta \mathbf{P}_i^n
\end{equation}
The intermediate poses $\{\mathbf{P}_i^j\}_{j=1}^{n-1}$ are calculated through linear interpolation, note that the interpolation of rotation is calculated on the $\mathfrak{so}(3)$ manifold \cite{sola2018micro}, implemented using PyPose \cite{wang2023pypose, zhan2023pypose}. The interpolation process is described as the following equations:
\begin{equation}
\mathbf{R}_i^j = \mathbf{R}_i^0 \cdot \exp (\frac{j}{n} \cdot \log((\mathbf{R}_i^0)^{-1} \cdot \mathbf{R}_i^n)),
\end{equation}
\begin{equation}
\mathbf{t}_i^j = \mathbf{t}_i^0 + \frac{j}{n} \cdot (\mathbf{t}_i^n - \mathbf{t}_i^0),
\end{equation}
where $\exp(\cdot)$ and $\log(\cdot)$ represents the exponential mapping and logarithmic mapping between $\mathbf{SO}(3)$ and $\mathfrak{so}(3)$.
During the evaluation process, we utilize the mid-point interpolation between $\mathbf{P}_i^0$ and $\mathbf{P}_i^{n}$, denoted as $\mathbf{P}_i^{mid}$ to serve as the optimized camera poses for rendering.

\subsubsection{Camera Motion Deblurring}
Provided with the set of virtual cameras, the corresponding blurry image can be synthesized as follows:
\begin{equation}
\widehat{\mathbf{B}}_i = \frac{1}{n+1}\sum^n_{j=0} \mathbf{I}_i^j(\mathbf{P}_i^j, \mathbf{G}_c^{static}),
\label{eq:blurry_reconstruction}
\end{equation}
where 
$\mathbf{I}_i^j(\mathbf{P}_i^j, \mathbf{G}_c^{static})$ denotes the virtual sharp image rendered with virtual pose $\mathbf{P}_i^j$ and the canonical space static Gaussians $\mathbf{G}_c^{static}$, $\widehat{\mathbf{B}}_i$ is the reconstructed blurry image. 

The loss function for learning virtual cameras as well as reconstructing the static region is a rendering loss between input blurry image $\mathbf{B}_i$ and reconstructed blurry image $\widehat{\mathbf{B}}_i$:
\begin{equation}
    \mathcal{L}_{render}  = \sum_{(x, y) \notin M_d} || \widehat{\mathbf{B}}_i(x, y) - \mathbf{B}_i(x, y) ||_1,
\label{eq:masked_rendering_loss}
\end{equation}
where $M_d$ is a dynamic mask obtained from Track-Anything \cite{yang2023track}, as we do not want the content within the dynamic region to affect the optimization of camera poses.

\subsection{Object Motion Modeling and Deblurring}
\label{subsec:dynamic_deblur}
After the camera motion deblurring stage, we removed the blurry effect caused by camera movement and reconstructed a sharp static region of the scene. In this section, we will dive into details on how we model object motion and apply object motion deblurring. 

\subsubsection{Object Motion Modeling}
\label{subsec:object_motion_modeling}
To model object motion, we utilize a multi-layer perceptron (MLP) to serve as a deformation field:
\begin{equation}
(\delta \mathbf{x}, \delta \mathbf{r}, \delta \mathbf{s})=\mathcal{F}_\theta(\gamma(\mathbf{x}), \gamma(t)),
\end{equation}
where $\mathcal{F}_\theta$ represents deformation field, $\gamma(\cdot)$ denotes positional encoding, $\mathbf{x}$ denotes the canonical space mean of the input dynamic Gaussian, $t$ is the input timestamp and $(\delta \mathbf{x}, \delta \mathbf{r}, \delta \mathbf{s})$ is the output offsets to transform canonical space Gaussians into the deformed space at time $t$.

Besides, to further refine the details in the static region, we introduce an extra deformation field to the static Gaussians. It can handle the byproducts resulting from the moving objects (\eg reflection and shadow) and improve the poor rendering quality in the marginal area due to the lack of initial Gaussians. In the following section, we denote the static deformation field and dynamic deformation field together as $\mathcal{F}_\theta$, for ease of understanding.

\subsubsection{Object Motion Deblurring}
\label{subsec:object_motion_deblurring}
Similar to camera motion deblurring in Sec.~\ref{subsec:static_deblur}, object motion deblurring is achieved by finding a set of virtual times $\{t_i^j\}_{j=0}^n$, where $t_i^j$ denotes the $j$th virtual timestamp around $t_i$, the corresponding timestamp of the blurry input $\mathbf{B}_i$. With these virtual times, we can query the deformation field to obtain the trajectories of Gaussians $\mathbf{G}_i^j$ (\ie virtual Gaussians) that represent the moving objects, where $\mathbf{G}_i^j = \mathbf{G}_c + \mathcal{F}_\theta(\gamma(\mathbf{X}), \gamma(t_i^j))$, in which $\mathbf{X}$ denotes the mean of the canonical space Gaussians, $\mathbf{G}_c$ is the canonical space Gaussians consists of $\mathbf{G}_c^{static}$ and $\mathbf{G}_c^{dyn}$. Notice that the number of virtual times should equal the number of virtual views. Thus, we can update Eq.~\ref{eq:blurry_reconstruction} to the following:
\begin{equation}
\widehat{\mathbf{B}}_i = \frac{1}{n+1}\sum^n_{j=0} \mathbf{I}_i^j(\mathbf{P}_i^j, \mathbf{G}_i^j),
\end{equation}
where $\mathbf{I}_i^j(\mathbf{P}_i^j, \mathbf{G}_i^j)$ is the virtual sharp image rendered with virtual camera pose $\mathbf{P}_i^j$ and the Gaussians $\mathbf{G}_i^j$ at virtual time $t_i^j$. We make an assumption that objects move at a constant speed within the exposure time, thus $\{t_i^j\}_{j=0}^n$ are interpolated uniformly in $[t_i-\frac{\tau}{2}, t_i+\frac{\tau}{2}]$, where $\tau$ denotes one exposure time. The training objective in this part is still the rendering loss, but considering the entire image:
\begin{equation}
    \mathcal{L}_{render}  = \sum_{(x, y) \in \mathbf{B}_i} || \widehat{\mathbf{B}}_i(x, y) - \mathbf{B}_i(x, y) ||_1.
\label{eq:rendering_loss}
\end{equation}

\subsubsection{Trajectory of Gaussians}
In the Subsec.~\ref{subsec:object_motion_deblurring}, we implicitly assume the deformation field will learn the actual trajectory of each Gaussian.
However, due to the implicit nature of the deformation field, we cannot constrain what Gaussian movements it will learn. According to our experiments, in most cases, Gaussians will not follow the movement of the objects, which means a single moving object will be represented by different groups of Gaussians at different times. Thus it cannot fulfill our hypothesis. To solve this issue, we introduce a 3D track loss to constrain the movement of the Gaussians at the initial stage of training, forcing them to follow the motion of 2D pixels:
\begin{equation}
    \mathcal{L}_{track} = \sum_{\mathbf{x}_i \in dynamic} || \mathbf{x}^{gt}_i - \mathbf{x}_i ||_1,
\label{eq:track_loss}
\end{equation}
where $\mathbf{x}_i$ is the learned dynamic Gaussian means produced by deformation field at time $i$, which is the time corresponds to the blurry input $\mathbf{B}_i$, and $\mathbf{x}^{gt}_i$ is the reference for dynamic Gaussian means.
We first utilize BootsTAPIR \cite{doersch2023tapir, doersch2024bootstap} to get a reference trajectory for each 2D pixel within the dynamic mask $M_d$. Next, we apply Depth-Anything \cite{depth_anything_v2} to obtain depth information. Then we utilize the optimized camera poses given in Sec.~\ref{subsubsec:camera_optimization} to backproject the 2D pixel trajectories into 3D space to form pseudo ground truth for 3D Gaussian trajectories $\{\mathbf{x}^{gt}_i\}_{i=1}^T$.
3D track loss not only encourages deformation field to learn the actual trajectories, it also helps model fast object motion better.


\begin{figure}
    \centering
    \includegraphics[width=0.99\linewidth]{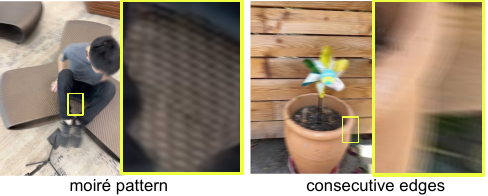} 
    \caption{\textbf{Demonstration of artifacts in synthesized dataset.}}
    \label{fig:synthesized_blur}
\end{figure}
\section{Dataset}
\label{sec:dataset}
\subsection{Synthesized Blurry Dataset}
\label{subsec:synthesized_data}
We generate a synthesized blurry dataset by following instructions in \cite{bui2023dyblurf} to create blurred images from clean scene images. Utilizing the pre-trained Video Frame Interpolation (VFI) model, RIFE \cite{huang2022rife}, we first increase the frame rate of the original input (e.g., from 30fps to 240fps). These interpolated sharp frames are then averaged to simulate the blur formation process, resulting in blurred images. This synthesized blurry dataset is used for deblurring evaluation

For this task, we select three scenes from the HyperNeRF dataset \cite{park2021hypernerf} and four scenes from the Dycheck dataset \cite{gao2022dynamic}, containing either relatively fast camera or object motion, to assess the effectiveness of our proposed method.

\subsection{Real-world Captured Blurry Dataset}
Though there exist several real-world blurry datasets \cite{li2022deblurnerf, Liu2021MBAVOMB} for 3D reconstruction, they lack dynamic objects in the scene for evaluating the performance of dynamic reconstruction.
Previous works \cite{bui2023dyblurf, sun2024dyblurf} utilize clean datasets to synthesize blurry datasets to evaluate the performance of their models. However, blurry data synthesized in this way (\ie Sec.~\ref{subsec:synthesized_data}) is often insufficiently blurred and may have some unrealistic artifacts, \ie moiré pattern and consecutive edges, as shown in Fig.~\ref{fig:synthesized_blur}. Also, due to the short exposure time and fast shutter speed of the camera, all the sharp frames were captured under good lighting conditions, which is not realistic for motion blurs to occur.

To solve the problem of lacking blurry dynamic scene data, we collect a real-world blurry dataset for dynamic reconstruction. Our dataset contains 12 scenes, each of which is captured with two GoPros. We set one of the GoPros to 1/24 second exposure time to capture motion blurred images for training and the other one is set to 1/240 second exposure time to capture sharp images for evaluation. The two GoPros are synchronized using timecode. 
We use this real-world blurry dataset to evaluate novel view synthesis. 
%

\begin{figure*}[!htbp]
    \centering
    \includegraphics[width=0.98\linewidth]{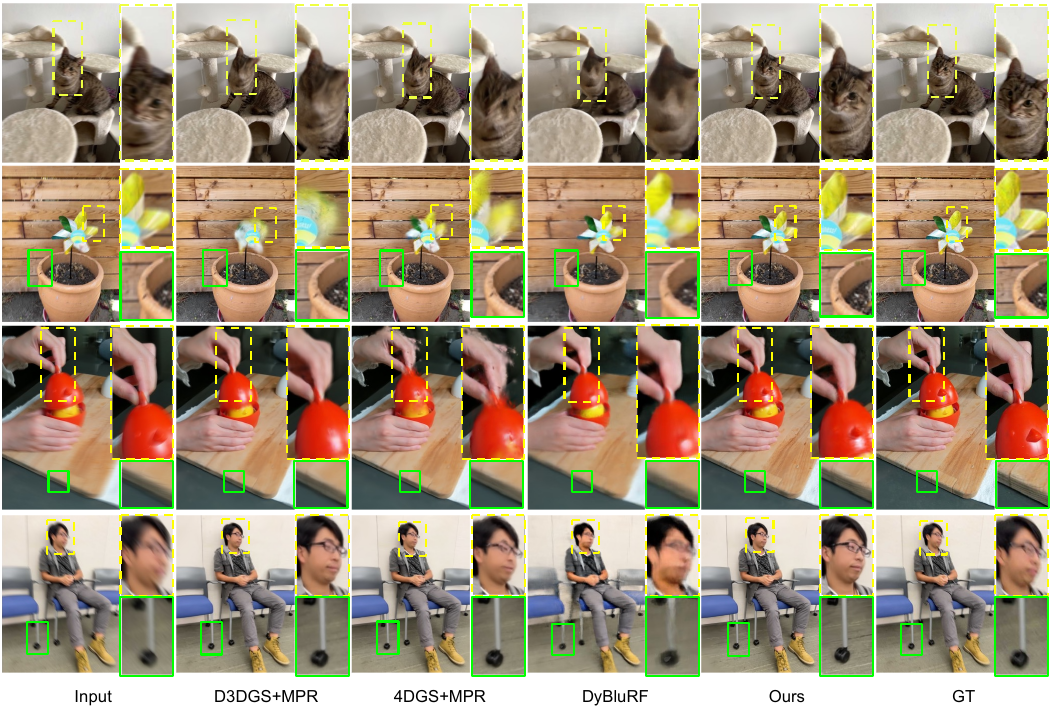}
    \vspace{-3mm}
    \caption{\textbf{Qualitative comparison of deblurring.} The \shadeText{green}{green} boxes represent details in static region and the dashed \shadeText{yellow}{yellow} boxes denote the dynamic part. As Dycheck dataset is not captured at high frame rate, our method produces a better result than GT in the last row.}
    \label{fig:deblurring}
\end{figure*}

\begin{table}[!tbp]
\centering
\caption{\textbf{Quantitative comparison of deblurring}, conducted on synthesized blurry dataset. Colors indicate the \shadeText{red}{best} and \shadeText{orange}{second best} results respectively. LV denotes Laplacian Variance.}
\resizebox{0.99\linewidth}{!}{ 
\begin{tabular}{c|ccccc}
\hline
\multicolumn{1}{l|}{\textbf{}} & \multicolumn{5}{c}{\textbf{Deblurring}}                                                                                                                          \\
Methods                        & PSNR↑                          & SSIM↑                          & LPIPS↓                         & LV↑                           & MUSIQ↑                        \\ \hline
DyBluRF \cite{sun2024dyblurf}                        & 23.82                          & 0.6900                         & 0.4714                         & 10.73                         & 32.95                         \\
D3DGS  \cite{yang2024deformable}                         & 27.20                          & 0.7986                         & 0.3486                         & 20.62                         & 37.74                         \\
D3DGS + \cite{Zamir2021MPRNet}                    & 27.98                          & \cellcolor[HTML]{F9CB9C}0.8372 & 0.3289                         & 36.34                         & 52.16                         \\
4DGS \cite{yang2023gs4d}                           & \cellcolor[HTML]{FFFFFF}27.39  & \cellcolor[HTML]{FFFFFF}0.8010 & \cellcolor[HTML]{FFFFFF}0.3399 & \cellcolor[HTML]{FFFFFF}18.48 & \cellcolor[HTML]{FFFFFF}39.14 \\
4DGS + \cite{Zamir2021MPRNet}                     & \cellcolor[HTML]{F9CB9C}28.17 & \cellcolor[HTML]{FFFFFF}0.8335 & \cellcolor[HTML]{F9CB9C}0.2641 & \cellcolor[HTML]{F9CB9C}47.74 & \cellcolor[HTML]{F9CB9C}54.00 \\ \hline
Ours                           & \cellcolor[HTML]{EA9999}28.75  & \cellcolor[HTML]{EA9999}0.8414 & \cellcolor[HTML]{EA9999}0.2343 & \cellcolor[HTML]{EA9999}65.31 & \cellcolor[HTML]{EA9999}57.03 \\ \hline
\end{tabular}
}
\label{tab:Deblurring}
\vspace{-3pt}
\end{table}

\begin{table}[!tbp]
\centering
\caption{\textbf{Quantitative comparison on real-world captured dataset.} When facing highly blurred images in our dataset, MPRNet fails to restore the sharp images. The experiment demonstrates our better performance compared with the baseline methods, including the pre-deblurred ones. LV denotes Laplacian Variance, and SI-* denotes shift-invariant metrics.}
\resizebox{0.99\linewidth}{!}{ 
\begin{tabular}{c|ccccc}
\hline
\multicolumn{1}{l|}{\textbf{}} & \multicolumn{5}{c}{\textbf{Novel View Synthesis}}                                                                                                               \\
Methods                        & SI-PSNR↑                      & SI-SSIM↑                       & SI-LPIPS↓                      & LV↑                           & MUSIQ↑                        \\ \hline
DyBluRF \cite{sun2024dyblurf}                        & 20.57                         & 0.7614                         & 0.3614                         & 35.16                         & 31.05                         \\
D3DGS \cite{yang2024deformable}                         & 22.83                         & 0.8243                         & 0.3500                         & 18.95                         & 28.89                         \\
D3DGS + \cite{Zamir2021MPRNet}                    & \cellcolor[HTML]{F9CB9C}23.10 & \cellcolor[HTML]{F9CB9C}0.8363 & \cellcolor[HTML]{F9CB9C}0.2637 & 37.58                         & 38.73                         \\
4DGS \cite{yang2023gs4d}                          & \cellcolor[HTML]{FFFFFF}22.06 & \cellcolor[HTML]{FFFFFF}0.8093 & \cellcolor[HTML]{FFFFFF}0.3733 & \cellcolor[HTML]{FFFFFF}24.60 & \cellcolor[HTML]{FFFFFF}30.17 \\
4DGS + \cite{Zamir2021MPRNet}                     & \cellcolor[HTML]{FFFFFF}22.58 & \cellcolor[HTML]{FFFFFF}0.8293 & \cellcolor[HTML]{FFFFFF}0.2863 & \cellcolor[HTML]{F9CB9C}50.68 & \cellcolor[HTML]{F9CB9C}41.88 \\ \hline
Ours                           & \cellcolor[HTML]{EA9999}25.13 & \cellcolor[HTML]{EA9999}0.8508 & \cellcolor[HTML]{EA9999}0.1567 & \cellcolor[HTML]{EA9999}96.42 & \cellcolor[HTML]{EA9999}54.29 \\ \hline
\end{tabular}
}
\label{tab:NVS}
\vspace{-6pt}
\end{table}

\section{Experiments}
\label{sec:exp}
\subsection{Implementation Details}
Our method is trained for 80,000 iterations. For the first 10,000 iterations, we use Eq.~\ref{eq:masked_rendering_loss} as the loss function to optimize virtual cameras as well as reconstruct the static region. Then for iteration 10,000 to 30,000, we introduce a static deformation field and use Eq.~\ref{eq:rendering_loss} as the rendering loss, together with Eq.~\ref{eq:track_loss} forcing the deformation field to learn from 2D pixel movements. 
Since we assume one-to-one correspondence between Gaussians and 3D track points, the splitting, duplicating and culling for dynamic Gaussians is temporarily suspended in this stage.
For the rest of the training process, we remove Eq.~\ref{eq:track_loss} and rely solely on Eq.~\ref{eq:rendering_loss} as the loss function, while resuming splitting, duplicating, and culling of dynamic Gaussians.

\subsection{Baselines and Evaluation Metrics}
\label{subsec:baseline_and_dataset}
\textbf{Baselines}.
We compare our method with DyBluRF \cite{sun2024dyblurf}, which is designed for dynamic reconstruction with motion blur and state-of-the-art dynamic reconstruction methods D3DGS \cite{yang2024deformable} and 4DGaussians \cite{yang2023gs4d} as well. 
Since D3DGS and 4DGaussians lack deblurring capabilities, we also pre-process the training images using the pre-trained deblurring method MPRNet \cite{Zamir2021MPRNet} to evaluate the performance.

\textbf{Evaluation Metrics}.
In addition to commonly used metrics like PSNR, SSIM, and LPIPS, we incorporate additional metrics to further showcase our rendering quality. Our experiments show that PSNR can sometimes yield higher values even when images appear blurrier. Therefore, we introduce Laplacian Variance (LV) and MUSIQ \cite{ke2021musiq} to better assess the sharpness and the overall quality of the rendered images. 
For Novel View Synthesis, we extract training poses and evaluation poses from blurry images and sharp images respectively. However, it is quite challenging to align these two sets of poses to the same local coordinate system perfectly, which will lead to a slight shift in the rendered image.
Thus, we utilize shift-invariant metrics in Tab. \ref{tab:NVS} for quantitative comparisons. For the shift-invariant metrics, we allow a 6-pixel shift in both horizontal and vertical axes, and choose the largest value among the 36 results.

\subsection{Deblurring}
For the deblurring task, we evaluate our method against baseline methods using synthesized datasets generated from both Dycheck \cite{gao2022dynamic}  and HyperNeRF \cite{park2021hypernerf}. The blurred images serve as input, and the original sharp images are used for evaluation.
Our quantitative comparison results are presented in Tab.~\ref{tab:Deblurring}. The results show that our method outperforms all baselines. The significant performance gap in LPIPS, LV, and MUSIQ scores indicate that DyBluRF, D3DGS, and 4DGS are suffering from motion-blurred inputs. 
Qualitative results are provided in Fig.~\ref{fig:deblurring}. 
Our method provides superior visual results in both static and dynamic regions. Notably, in the last row, our rendering results appear even clearer than the ground truth. This is because the ground truth images in the Dycheck dataset were captured at a low frame rate, resulting in some blurring.

\begin{figure*}
    \centering
    \includegraphics[width=0.98\linewidth]{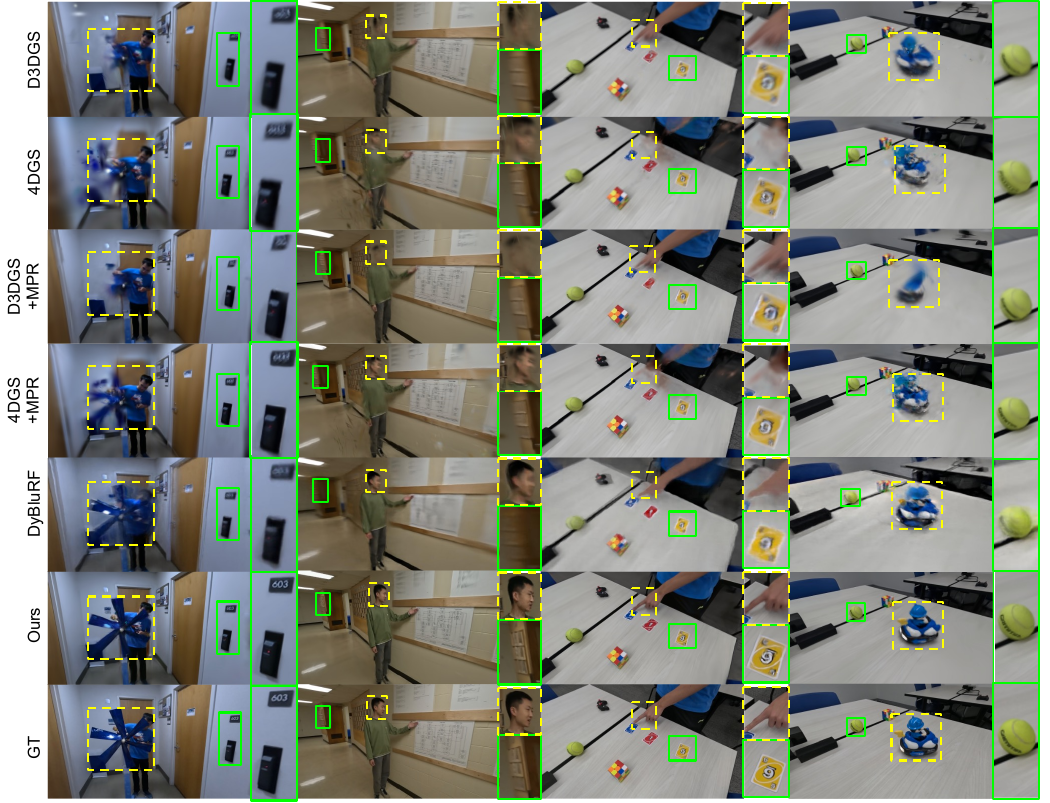}
    \caption{\textbf{Qualitative comparison of Novel View Synthesis.} Our method shows a significantly better rendering quality in the dynamic region, \eg the windmill, the face details, card and the toy car. Besides, for static region, our method reconstructs finer details and shows a even better result than the ground truth image, \eg the details in the green bounding box in the second column.}
    \label{fig:NVS}
\end{figure*}
\begin{table*}[!htbp]
\centering
\caption{\textbf{Ablation studies on each module of our proposed method.} TL: Track Loss, DB: Dynamic Deblurring, SD: Static Deblurring, SDF: Static Deformation Field. LV denotes Laplacian Variance, and SI-* denotes shift-invariant metrics.}
\resizebox{0.94\linewidth}{!}{ 
\begin{tabular}{c|ccccc|ccccc}
\hline
\multicolumn{1}{l|}{\textbf{}} & \multicolumn{5}{c|}{\textbf{Deblurring}}                                             & \multicolumn{5}{c}{\textbf{Novel View Synthesis}}                                    \\
\multicolumn{1}{l|}{}          & PSNR↑          & SSIM↑           & LPIPS↓          & LV↑            & MUSIQ↑         & SI-PSNR↑       & SI-SSIM↑        & SI-LPIPS↓       & LV↑            & MUSIQ↑         \\ \hline
w/o TL                         & 27.68          & 0.8186          & 0.2757          & 57.53          & 54.69          & 24.12          & 0.8272          & 0.1993          & 88.24          & 50.35          \\
w/o DB                         & 28.47          & 0.8229          & 0.2614          & 59.31          & 55.59          & 24.11          & 0.8229          & 0.1873          & 91.76          & 51.87          \\
w/o SD                         & 26.00          & 0.7514          & 0.3743          & 16.42          & 36.57          & 22.22          & 0.7656          & 0.2702          & 24.69          & 29.51          \\
w/o SDF                        & 27.19          & 0.8057          & 0.2814          & 59.42          & 52.37          & 24.40          & 0.8244          & 0.2534          & 76.58          & 44.05          \\
\textbf{Ours}                   & \textbf{28.75} & \textbf{0.8414} & \textbf{0.2343} & \textbf{65.31} & \textbf{57.03} & \textbf{25.13} & \textbf{0.8508} & \textbf{0.1567} & \textbf{96.42} & \textbf{54.29} \\ \hline
\end{tabular}
}
\vspace{-3mm}
\label{tab:ablation_component}
\end{table*}
\begin{table}[!htbp]
\Large
\centering
\caption{\textbf{Ablation studies on the number of virtual views.} Each color indicates the \shadeText{red}{best}, \shadeText{orange}{second best} and \shadeText{yellow}{third best} respectively. The results indicate that increasing virtual cameras enhances model performance but peaks at a certain point.}
\resizebox{0.999\linewidth}{!}{ 
\renewcommand{\arraystretch}{1.1} 
\begin{tabular}{c|ccc|ccc}
\hline
\multicolumn{1}{l|}{\textbf{}}        & \multicolumn{3}{c|}{\textbf{toycar}}                                                            & \multicolumn{3}{c}{\textbf{windmill}}                                                           \\
\multicolumn{1}{c|}{\# virtual}       & \multirow{2}{*}{SI-PSNR↑}     & \multirow{2}{*}{SI-SSIM↑}      & \multirow{2}{*}{SI-LPIPS↓}     & \multirow{2}{*}{SI-PSNR↑}     & \multirow{2}{*}{SI-SSIM↑}      & \multirow{2}{*}{SI-LPIPS↓}     \\
\multicolumn{1}{c|}{views}            &                                &                                &                                &                                &                                &                                \\ \hline
6                                     & \cellcolor[HTML]{F9CB9C}23.45 & \cellcolor[HTML]{FFFC9E}0.8660 & \cellcolor[HTML]{FFFC9E}0.1557 & 26.74                         & \cellcolor[HTML]{FFFC9E}0.9034 & \cellcolor[HTML]{FFFC9E}0.1421 \\
8                                     & \cellcolor[HTML]{EA9999}23.63 & \cellcolor[HTML]{F9CB9C}0.8665 & \cellcolor[HTML]{F9CB9C}0.1546 & \cellcolor[HTML]{FFFC9E}26.75 & 0.9028                         & 0.1424                         \\
10                                    & \cellcolor[HTML]{FFFC9E}23.47 & \cellcolor[HTML]{EA9999}0.8676 & \cellcolor[HTML]{EA9999}0.1419 & \cellcolor[HTML]{EA9999}27.52 & \cellcolor[HTML]{EA9999}0.9063 & \cellcolor[HTML]{EA9999}0.1275 \\
15                                    & 22.96                         & 0.8532                         & 0.1667                         & \cellcolor[HTML]{F9CB9C}26.83 & \cellcolor[HTML]{F9CB9C}0.9062 & \cellcolor[HTML]{F9CB9C}0.1376 \\
20                                    & 23.33                         & 0.8625                         & 0.1589                         & 26.31                         & 0.8977                         & 0.1388                         \\ \hline
\end{tabular}
}

\vspace{-10pt} 

\label{tab:ablation_virtual_view}
\end{table}

\subsection{Novel View Synthesis}
For the novel view synthesis task, we compare our method with baseline approaches on our real-world blurry dataset, as stated in Sec.~\ref{subsec:baseline_and_dataset}. 
The quantitative results are shown in Tab.~\ref{tab:NVS}. 
When facing highly blurred images in our dataset, MPRNet fails to restore the sharp images.
Our method delivers superior results across all aspects, with a wider performance gap compared to the deblurring task.
We also provide a qualitative comparison in Fig.~\ref{fig:NVS}. 
Our method demonstrates significantly better rendering quality within dynamic regions while preserving fine details in static areas.

\subsection{Ablation Study}
\vspace{-2pt}
\textbf{Component ablation.}
We first conduct ablation studies to evaluate the contributions of each component in our proposed method on both deblurring and novel view synthesis tasks. We evaluate the impact of (1) removing tracking loss (w/o TL), (2) removing dynamic deblurring (w/o DB), (3) removing static deblurring (w/o SD) and (4) removing static deformation field (w/o SDF). The quantitative results are shown in Tab.~\ref{tab:ablation_component}. We can observe a huge performance degradation without static deblurring, which is reasonable, since static region dominates the scene. Removing either tracking loss or dynamic deblurring will result in a lower rendering quality. One thing to be noticed is that removing dynamic deblurring in the deblurring task has less impact because the scenes in the synthesized dataset generally have less blurry effect than the real-world blurry dataset.
\\
\textbf{Virtual camera poses.}
We also conduct experiments to show the effect of the number of virtual views within the exposure time, as described in Sec.~\ref{subsec:static_deblur}. We choose a set of numbers ranging from 6 to 20, representing the number of virtual views, and randomly pick two scenes (\ie toycar and windmill) to conduct this ablation study. The quantitative results are shown in Tab.~~\ref{tab:ablation_virtual_view}. Our results indicate that increasing the number of virtual views will help improves motion blur handling up to a certain point, after which additional views become counterproductive.

\section{Conclusion}
\label{conclusion}
We introduce BARD-GS, a Gaussian Splatting-based method for dynamic scene reconstruction from blurry inputs. By explicitly decoupling and modeling both object and camera motion blur, our approach achieves superior reconstruction quality compared to previous methods. In addition, we present a real-world blurry dataset for dynamic scene reconstruction, which will be released to benefit further research in the community. 

{
    \small
    \bibliographystyle{ieeenat_fullname}
    \bibliography{main}
}
\clearpage
\setcounter{page}{1}
\maketitlesupplementary

\section{Real-world Motion Blur Dataset}
\label{sec:dataset}

Motion blur is a common occurrence in everyday photography, especially in dynamic scenes. Our proposed dataset addresses a crucial gap in existing resources by capturing realistic blurs caused by camera motion and fast-moving objects, making it highly relevant for studying real-world dynamic scene reconstruction. Table~\ref{tab:dataset_comparison} compares the key characteristics across different datasets. While most datasets either lack blur entirely (\eg HyperNeRF~\cite{park2021hypernerf} and iPhone~\cite{gao2022dynamic}) or rely on synthetic blur (\eg Stereo Blur~\cite{Zhou2019Stereodeblur}), ours and Deblur-NeRF~\cite{ma2022deblur} are the only ones incorporating real-world blur. However, unlike Deblur-NeRF, which focuses on static scenes without object motion blur, our dataset offers a comprehensive and realistic benchmark for evaluating and developing solutions to complex motion blur challenges in dynamic settings.

\subsection{Data Acquisition Process}

\begin{figure*}
    \centering
    \includegraphics[width=1\linewidth]{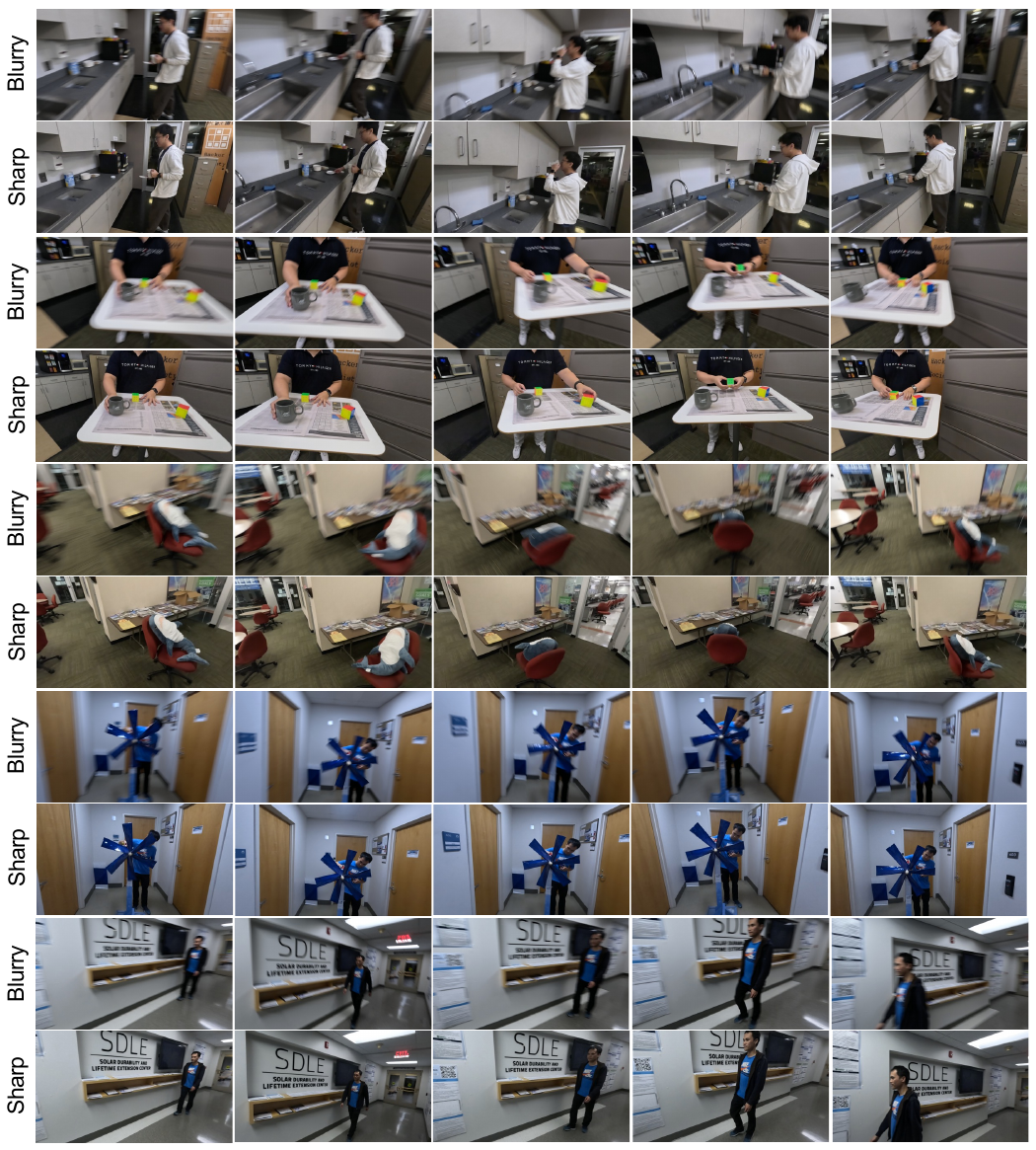}
    \caption{\textbf{Samples from our real-world blurry dataset.} Each row pair showcases blurry input frames (top of each pair) alongside their corresponding sharp ground-truth frames (bottom of each pair). The dataset captures diverse scenarios with fast-moving objects and camera motion, reflecting real-world challenges.}
    \label{fig:dataset}
\end{figure*}

We present the first real-world motion blur dataset specifically designed for dynamic scene reconstruction. The captured videos incorporate camera motions and fast-moving objects, which are typically encountered in real-world scenarios. Figure~\ref{fig:dataset} illustrates sample frames from our dataset. 

Our dataset is captured using two synchronized GoPro Hero 12 cameras. One camera is configured with a 1/24-second shutter speed to record 24 FPS \textit{blurry} videos, while the other employs a 1/240-second shutter speed to capture 240 FPS \textit{sharp} videos. The sharp videos are used for evaluation purposes only. To ensure precise synchronization, we utilized the timecode feature\footnote{https://gopro.github.io/labs/control/precisiontime/} provided by GoPro Lab firmware, complemented by manual timestamp alignment. This approach achieves a time offset of less than 1 millisecond, ensuring high temporal accuracy.

\subsection{Data Pre-processing}
\textbf{Camera Pose Estimation.} We begin by running COLMAP on all sharp images to obtain accurate camera poses, which serve as a reference for evaluation. Blurry images are then sequentially registered to this COLMAP reconstruction, producing intentionally inaccurate camera poses for training. This process aligns both sharp and blurry images within a unified world space, enabling precise evaluations.
Additionally, the intentional inaccuracy of the blurry image camera poses mirrors real-world conditions, enhancing the realism of the training setup.

\noindent\textbf{Point Cloud Initialization.} Using these intentionally inaccurate camera poses, we perform a second reconstruction to generate an initial point cloud for training. 
Our method refines and optimizes the camera poses, resulting in final reconstructions that may exhibit slight shifts compared to those based on the unoptimized poses. Since other methods do not update poses, we ensure fair comparisons during evaluation by using shift-invariant metrics that tolerate minor pixel-level shifts in rendered images.

\begin{figure}
    \centering
    \includegraphics[width=1\linewidth]{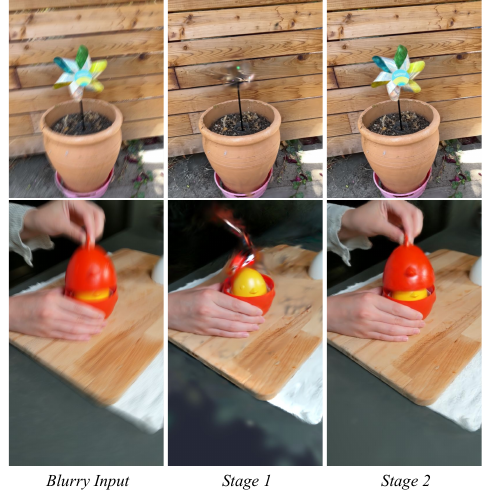}
    \caption{\textbf{Illustration of the two-stage training process.} The 2$^{nd}$ column shows outputs of \textit{Stage 1}, where static regions are sharply reconstructed (\eg, the wooden background, flowerpot (top row), and cutting board (bottom row)). The 3$^{rd}$ column displays the outputs of \textit{Stage 2}, where dynamic components, such as the spinning windmill (top row) and the lifting cap (bottom row), are clarified.}
    \label{fig:2stage}
\end{figure}

\begin{table*}
\centering
\caption{\textbf{Comparison of Dataset Characteristics.} Our dataset bridges a critical gap in current benchmarks by capturing real-world motion blur caused by both camera movement and fast-moving objects in dynamic scenes, providing a realistic and advanced resource for research.}
\resizebox{0.999\linewidth}{!}{ 
\begin{tabular}{cccccccc}
\toprule
\multirow{2}{*}{Dataset} & \multirow{2}{*}{Dynamic Scene} & \multirow{2}{*}{Blur Type} & \multicolumn{2}{c}{Blur Cause} & \multirow{2}{*}{\# Cameras} & \multirow{2}{*}{\# Scenes} & \multirow{2}{*}{Frame Rate (FPS)}  \\ 
        \cline{4-5} 
        &&& Camera Motion & Object Motion && \\
\toprule
HyperNeRF~\cite{park2021hypernerf} & \checkmark  & $\times$  & $\times$ & $\times$ & 2 & 17 & 15 \\
iPhone~\cite{gao2022dynamic} & \checkmark & $\times$ & $\times$ & $\times$ & 3 & 14 & 30 / 60 \\
Deblur-NeRF~\cite{ma2022deblur} & $\times$ & Real-World  & \checkmark & $\times$ & 1 & 30 & -- \\
Stereo Blur~\cite{Zhou2019Stereodeblur} & \checkmark  & Synthetic  & \checkmark  & Slow & 2 & 135 & 60 \\ 
\hline
Ours & \checkmark & Real-World & \checkmark & Fast & 2 & 12 & 24 / 240  \\ 
\bottomrule
\end{tabular}
}
\label{tab:dataset_comparison}
\end{table*}

\section{Two-Stage Training}
Figure~\ref{fig:2stage} illustrates the outputs at each stage of our proposed framework, which progressively refines the static and dynamic components within the scene.

In \textit{stage 1}, we focus on mitigating camera motion blur by optimizing the camera trajectory, resulting in sharp reconstructions of static regions (\eg, the wooden background, floor, and flowerpot in the top-row example, and the table and cutting board in the bottom-row example). Since supervision is limited to the static regions during this stage, the reconstructions of dynamic regions remain incomplete and exhibit random artifacts.

In \textit{stage 2}, we address object motion blur by modeling object movements using Gaussian motion trajectories. This stage allows for clearer reconstruction of dynamic regions, such as the spinning windmill in the top row and the cap being lifted in the bottom row. Simultaneously, static details are further refined through the introduction of a deformation field for static Gaussians, ensuring the overall enhancement of scene details.

\section{Additional Results of Novel View Synthesis}
\begin{figure*}
    \centering
    \includegraphics[width=1\linewidth]{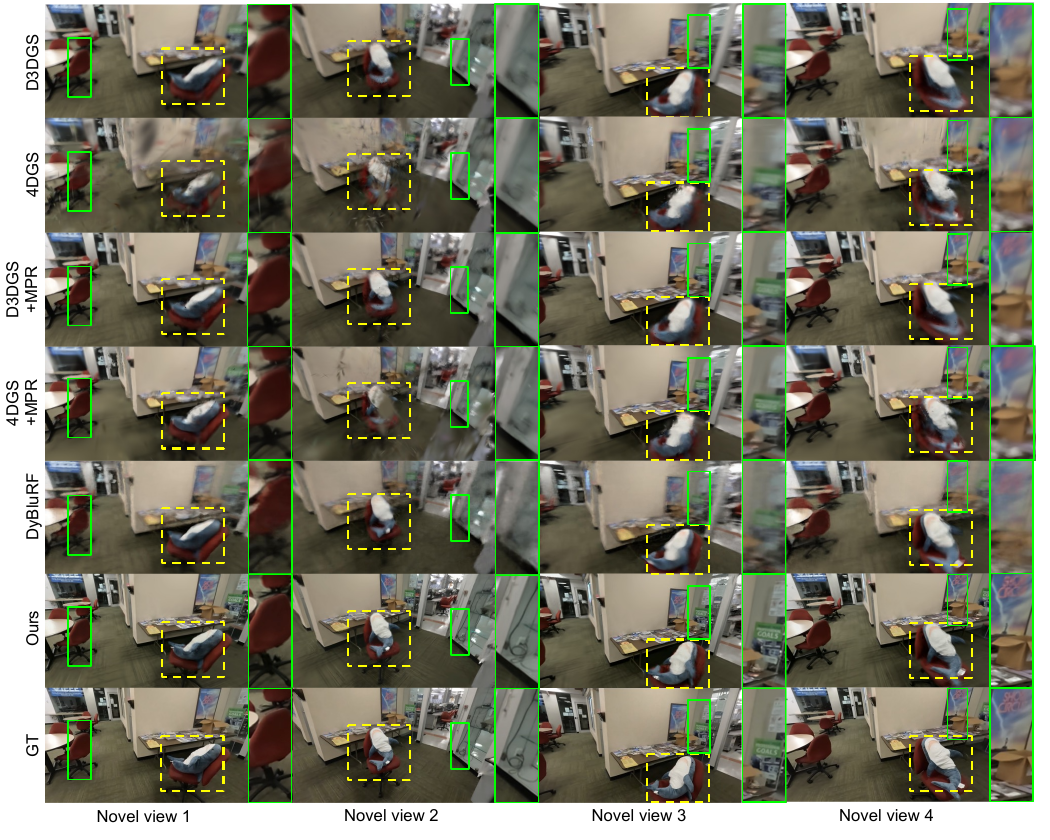}
    \caption{\textbf{Novel view comparison on the \textit{``shark-spin"} scene.} \shadeText{green}{Green} boxes highlight details in static regions, while the dashed \shadeText{yellow}{yellow} boxes indicate dynamic areas. Our method demonstrates significantly better reconstruction in the dynamic region (\ie, the shark toy on the spinning chair) and preserves superior details in static regions (\eg, the red chair in the background, wires in the back, and posters on the wall) across all novel views.}
    \label{fig:novel_view_1}
\end{figure*}

\begin{figure*}
    \centering
    \includegraphics[width=1\linewidth]{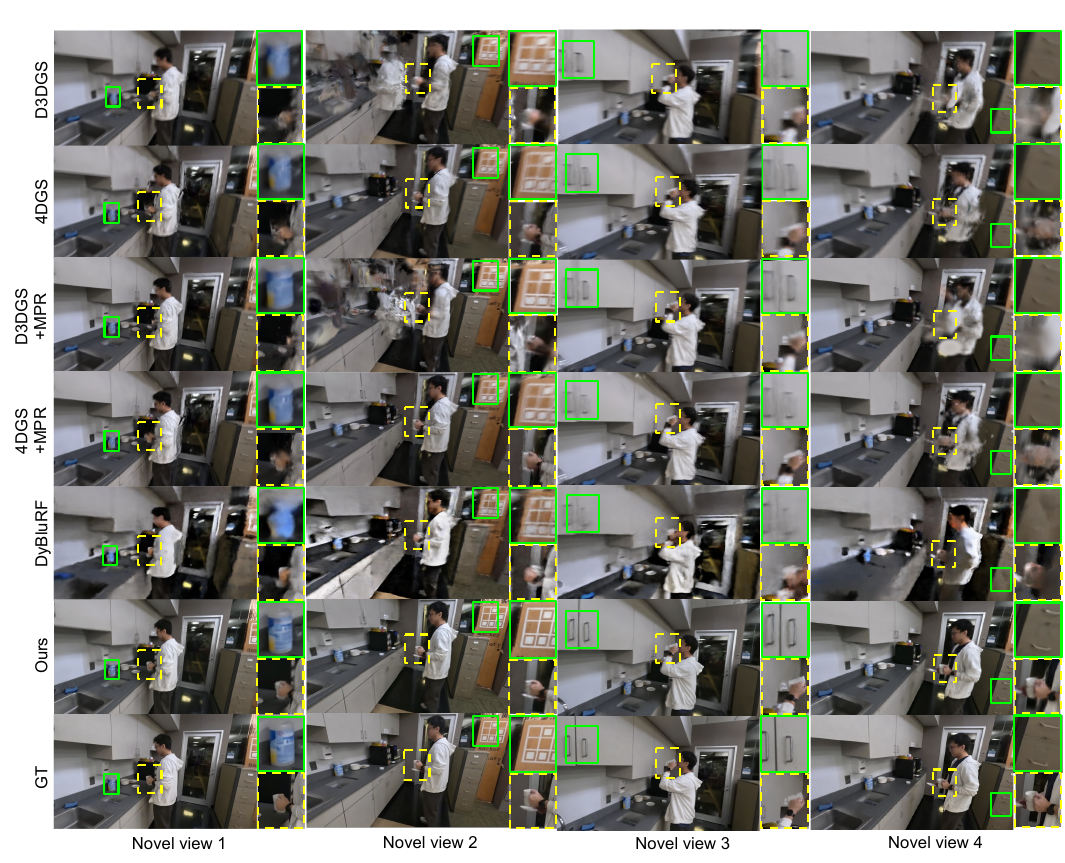}
    \caption{\textbf{Novel view comparison on the \textit{``kitchen"} scene.} \shadeText{green}{Green} boxes highlight fine details in static regions, and the dashed \shadeText{yellow}{yellow} boxes indicate dynamic parts. Our method demonstrates superior reconstruction in static regions, accurately capturing intricate elements such as the cleaning bottle, wall painting, and cabinet handles. Additionally, it effectively mitigates blur caused by moving objects, such as the moving hand and body. These results emphasize our method's ability to preserve detailed textures and address challenging visual artifacts in both static and dynamic regions.}
    \label{fig:novel_view_2}
\end{figure*}

To further demonstrate the effectiveness of our proposed method, we show additional comparison results for novel view synthesis in Figure~\ref{fig:novel_view_1} and~\ref{fig:novel_view_2}. Each scene includes four distinct novel views, with dynamic regions highlighted by dashed yellow boxes and static regions by green boxes. Our method consistently outperforms baseline approaches, delivering superior reconstruction quality in both static and dynamic regions.

Additionally, quantitative comparisons of novel view synthesis across 7 captured scenes in our dataset are provided in  Table~\ref{tab:full_NVS}. We report the results of five metrics, including shift-invariant PSNR (SI-PSNR), shift-invariant SSIM~\cite{wang2004image} (SI-SSIM), shift-invariant LPIPS~\cite{zhang2018unreasonable} (SI-LPIPS), Laplacian Variance score (LV)~\cite{bansal2016blur}, and MUSIQ~\cite{ke2021musiq}.
Our approach achieves the best performance for the majority of scenes across all five metrics, underscoring its robustness and effectiveness.

\section{Ablation Study of Proposed Components}
\begin{figure*}
    \centering
    \includegraphics[width=1\linewidth]{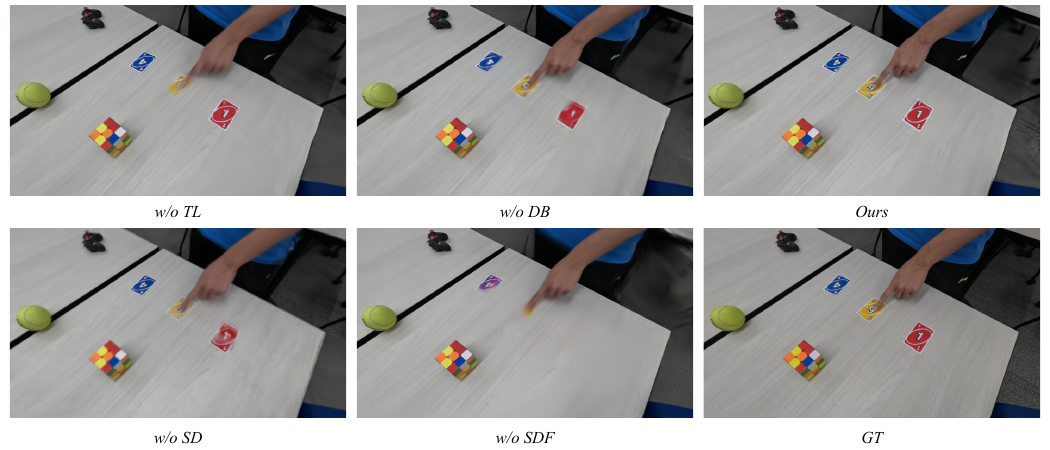}
    \caption{\textbf{Visual comparison of component ablation study.} \textit{w/o TL} denotes the exclusion of tracking loss ($\mathcal{L}_{track}$), while \textit{w/o DB} indicates the removal of \textit{stage 2}: \underline{\textit{D}}ynamic region de\underline{\textit{B}}lurring. Similarly, \textit{w/o SD} refers to the absence of \textit{stage 1}: \underline{\textit{S}}tatic region \underline{\textit{D}}eblurring, and \textit{w/o SDF} signifies the exclusion of \underline{\textit{S}}tatic \underline{\textit{D}}eformation \underline{\textit{F}}ield in \textit{stage 2}. These comparisons highlight the critical contributions of each component of the proposed approach to achieving clear and artifact-free results.}
    \label{fig:component_ablation}
\end{figure*}

In this section, we illustrate the impact of each component on the final reconstruction quality, as shown in Figure~\ref{fig:component_ablation}. 
We can see that removing either tracking loss (\textit{w/o TL}) or object motion deblurring (\textit{w/o DB}) results in significant blurring of dynamic regions. Omitting static deblurring (\textit{w/o SD}) causes blurring in the static regions, while the exclusion of static deformation field (\textit{w/o SDF}) introduces floating artifacts in the static regions, leading to occlusions and distortions in the dynamic regions. These results emphasize the importance of each component in our method for achieving clear, artifact-free results.

\begin{table*}[]
\Large
\renewcommand{\arraystretch}{1}
\centering
\caption{\textbf{Quantitative novel view synthesis comparison across 7 captured scenes in our dataset.} The results are evaluated using five metrics: SI-PSNR (↑), SI-SSIM (↑), SI-LPIPS (↓), LV (↑), and MUSIQ (↑). Colors indicate the \shadeText{red}{best} and \shadeText{orange}{second best} results, respectively. Our approach achieves the best results across all scenes for all five metrics in most cases.}
\resizebox{1\linewidth}{!}{ 
\begin{tabular}{c|ccccc|cccccccccc}
\hline
\multicolumn{1}{l|}{\textbf{}} & \multicolumn{5}{c|}{\textbf{card}}                                                                                                                              & \multicolumn{5}{c|}{\textbf{toycar}}                                                                                                                                                  & \multicolumn{5}{c}{\textbf{windmill}}                                                                                                                           \\
\multicolumn{1}{l|}{}          & SI-PSNR↑                      & SI-SSIM↑                       & SI-LPIPS↓                      & LV↑                           & MUSIQ↑                        & SI-PSNR↑                      & SI-SSIM↑                       & SI-LPIPS↓                      & LV↑                            & \multicolumn{1}{c|}{MUSIQ↑}                        & SI-PSNR↑                      & SI-SSIM↑                       & SI-LPIPS↓                      & LV↑                           & MUSIQ↑                        \\ \hline
DyBluRF                        & 19.86                         & 0.7602                         & 0.4102                         & 13.75                         & 35.11                         & 18.15                         & 0.7912                         & 0.3303                         & 41.68                          & \multicolumn{1}{c|}{36.96}                         & 21.79                         & 0.7932                         & 0.2630                         & 14.94                         & 33.32                         \\
D3DGS                          & 23.97                         & 0.8323                         & 0.3111                         & 19.45                         & 37.02                         & 20.58                         & 0.8588                         & 0.2811                         & 29.51                          & \multicolumn{1}{c|}{43.33}                         & 20.32                         & 0.8014                         & 0.4696                         & 8.06                          & 25.25                         \\
D3DGS+MPR                      & \cellcolor[HTML]{F9CB9C}24.09 & 0.8323                         & \cellcolor[HTML]{F9CB9C}0.2226 & 59.69                         & 50.54                         & 20.06                         & 0.8529                         & 0.3213                         & 37.12                          & \multicolumn{1}{c|}{43.53}                         & 22.63                         & \cellcolor[HTML]{F9CB9C}0.8632 & \cellcolor[HTML]{F9CB9C}0.2401 & 15.37                         & 34.21                         \\
4DGS                           & 22.33                         & 0.8290                         & 0.3323                         & 38.46                         & 40.09                         & 20.84                         & 0.8519                         & 0.2935                         & 35.72                          & \multicolumn{1}{c|}{43.75}                         & 21.37                         & 0.8236                         & 0.4022                         & 8.76                          & 24.47                         \\
4DGS+MPR                       & 22.58                         & \cellcolor[HTML]{EA9999}0.8459 & 0.2809                         & \cellcolor[HTML]{F9CB9C}76.38 & \cellcolor[HTML]{F9CB9C}54.29 & \cellcolor[HTML]{F9CB9C}20.94 & \cellcolor[HTML]{F9CB9C}0.8621 & \cellcolor[HTML]{F9CB9C}0.2513 & \cellcolor[HTML]{EA9999}74.51  & \multicolumn{1}{c|}{\cellcolor[HTML]{F9CB9C}55.07} & \cellcolor[HTML]{F9CB9C}23.29 & 0.8575                         & 0.2855                         & \cellcolor[HTML]{F9CB9C}20.10 & \cellcolor[HTML]{F9CB9C}36.40 \\
Ours                           & \cellcolor[HTML]{EA9999}25.76 & \cellcolor[HTML]{F9CB9C}0.8332 & \cellcolor[HTML]{EA9999}0.1701 & \cellcolor[HTML]{EA9999}78.29 & \cellcolor[HTML]{EA9999}63.38 & \cellcolor[HTML]{EA9999}23.47 & \cellcolor[HTML]{EA9999}0.8676 & \cellcolor[HTML]{EA9999}0.1419 & \cellcolor[HTML]{F9CB9C}60.04  & \multicolumn{1}{c|}{\cellcolor[HTML]{EA9999}57.17} & \cellcolor[HTML]{EA9999}27.52 & \cellcolor[HTML]{EA9999}0.9063 & \cellcolor[HTML]{EA9999}0.1275 & \cellcolor[HTML]{EA9999}69.11 & \cellcolor[HTML]{EA9999}48.44 \\ \hline
\multicolumn{1}{l|}{\textbf{}} & \multicolumn{5}{c|}{\textbf{shark-spin}}                                                                                                                        & \multicolumn{5}{c|}{\textbf{walk}}                                                                                                                                                    & \multicolumn{5}{c}{\textbf{kitchen}}                                                                                                                            \\
\multicolumn{1}{l|}{}          & SI-PSNR↑                      & SI-SSIM↑                       & SI-LPIPS↓                      & LV↑                           & MUSIQ↑                        & SI-PSNR↑                      & SI-SSIM↑                       & SI-LPIPS↓                      & LV↑                            & \multicolumn{1}{c|}{MUSIQ↑}                        & SI-PSNR↑                      & SI-SSIM↑                       & SI-LPIPS↓                      & LV↑                           & MUSIQ↑                        \\ \hline
DyBluRF                        & 22.59                         & 0.7816                         & 0.4099                         & 11.39                         & 18.14                         & 19.34                         & 0.7301                         & 0.2732                         & \cellcolor[HTML]{F9CB9C}123.51 & \multicolumn{1}{c|}{40.30}                         & 18.76                         & 0.7004                         & 0.4311                         & \cellcolor[HTML]{F9CB9C}27.83 & 23.71                         \\
D3DGS                          & \cellcolor[HTML]{F9CB9C}24.09 & 0.8192                         & 0.3814                         & 11.43                         & 19.77                         & \cellcolor[HTML]{F9CB9C}21.99 & 0.7880                         & 0.3067                         & 31.81                          & \multicolumn{1}{c|}{24.76}                         & 22.87                         & 0.8223                         & 0.3798                         & 10.89                         & 23.85                         \\
D3DGS+MPR                      & 24.06                         & \cellcolor[HTML]{F9CB9C}0.8369 & \cellcolor[HTML]{F9CB9C}0.3156 & 17.73                         & 28.74                         & 21.68                         & \cellcolor[HTML]{F9CB9C}0.7965 & \cellcolor[HTML]{F9CB9C}0.2165 & 80.01                          & \multicolumn{1}{c|}{41.22}                         & \cellcolor[HTML]{F9CB9C}22.94 & \cellcolor[HTML]{EA9999}0.8424 & 0.2690                         & 18.26                         & 31.21                         \\
4DGS                           & 21.23                         & 0.7676                         & 0.4922                         & 12.16                         & 21.39                         & 20.94                         & 0.7641                         & 0.3247                         & 47.37                          & \multicolumn{1}{c|}{27.59}                         & 22.66                         & 0.8109                         & 0.3523                         & 14.44                         & 24.64                         \\
4DGS+MPR                       & 22.33                         & 0.8011                         & 0.3739                         & \cellcolor[HTML]{F9CB9C}23.16 & \cellcolor[HTML]{F9CB9C}28.86 & 20.79                         & 0.773                          & 0.2529                         & 100.26                         & \multicolumn{1}{c|}{\cellcolor[HTML]{F9CB9C}45.51} & 22.81                         & 0.8325                         & \cellcolor[HTML]{F9CB9C}0.2460 & 27.35                         & \cellcolor[HTML]{F9CB9C}31.34 \\
Ours                           & \cellcolor[HTML]{EA9999}25.43 & \cellcolor[HTML]{EA9999}0.8562 & \cellcolor[HTML]{EA9999}0.1575 & \cellcolor[HTML]{EA9999}96.45 & \cellcolor[HTML]{EA9999}49.88 & \cellcolor[HTML]{EA9999}23.31 & \cellcolor[HTML]{EA9999}0.8268 & \cellcolor[HTML]{EA9999}0.1440 & \cellcolor[HTML]{EA9999}216.54 & \multicolumn{1}{c|}{\cellcolor[HTML]{EA9999}52.05} & \cellcolor[HTML]{EA9999}23.47 & \cellcolor[HTML]{F9CB9C}0.8326 & \cellcolor[HTML]{EA9999}0.1670 & \cellcolor[HTML]{EA9999}68.84 & \cellcolor[HTML]{EA9999}45.32 \\ \hline
\multicolumn{1}{l|}{\textbf{}} & \multicolumn{5}{c|}{\textbf{poster}}                                                                                                                            & \multicolumn{5}{c}{\textbf{average}}                                                                                                                                                  & \multicolumn{1}{l}{}          & \multicolumn{1}{l}{}           & \multicolumn{1}{l}{}           & \multicolumn{1}{l}{}          & \multicolumn{1}{l}{}          \\
\multicolumn{1}{l|}{}          & SI-PSNR↑                      & SI-SSIM↑                       & SI-LPIPS↓                      & LV↑                           & MUSIQ↑                        & SI-PSNR↑                      & SI-SSIM↑                       & SI-LPIPS↓                      & LV↑                            & MUSIQ↑                                             & \multicolumn{1}{l}{}          & \multicolumn{1}{l}{}           & \multicolumn{1}{l}{}           & \multicolumn{1}{l}{}          & \multicolumn{1}{l}{}          \\ \cline{1-11}
DyBluRF                        & 23.51                         & 0.7803                         & 0.4199                         & 13.03                         & 29.84                         & 20.57                         & 0.7614                         & 0.3614                         & 35.16                          & 31.05                                              & \multicolumn{1}{l}{}          & \multicolumn{1}{l}{}           & \multicolumn{1}{l}{}           & \multicolumn{1}{l}{}          & \multicolumn{1}{l}{}          \\
D3DGS                          & 25.99                         & \cellcolor[HTML]{EA9999}0.8396 & 0.3280                         & 21.50                         & 28.25                         & 22.83                         & 0.8243                         & 0.3500                         & 18.95                          & 28.89                                              & \multicolumn{1}{l}{}          & \multicolumn{1}{l}{}           & \multicolumn{1}{l}{}           & \multicolumn{1}{l}{}          & \multicolumn{1}{l}{}          \\
D3DGS+MPR                      & \cellcolor[HTML]{F9CB9C}26.22 & 0.8302                         & \cellcolor[HTML]{F9CB9C}0.2607 & \cellcolor[HTML]{F9CB9C}34.87 & \cellcolor[HTML]{F9CB9C}41.64 & \cellcolor[HTML]{F9CB9C}23.10 & \cellcolor[HTML]{F9CB9C}0.8363 & \cellcolor[HTML]{F9CB9C}0.2637 & 37.58                          & 38.73                                              & \multicolumn{1}{l}{}          & \multicolumn{1}{l}{}           & \multicolumn{1}{l}{}           & \multicolumn{1}{l}{}          & \multicolumn{1}{l}{}          \\
4DGS                           & 24.99                         & 0.8177                         & 0.4160                         & 15.27                         & 29.25                         & 22.06                         & 0.8093                         & 0.3733                         & 24.60                          & 30.17                                              & \multicolumn{1}{l}{}          & \multicolumn{1}{l}{}           & \multicolumn{1}{l}{}           & \multicolumn{1}{l}{}          & \multicolumn{1}{l}{}          \\
4DGS+MPR                       & 25.30                         & 0.8332                         & 0.3135                         & 32.93                         & 41.63                         & 22.58                         & 0.8293                         & 0.2863                         & \cellcolor[HTML]{F9CB9C}50.68  & \cellcolor[HTML]{F9CB9C}41.88                      & \multicolumn{1}{l}{}          & \multicolumn{1}{l}{}           & \multicolumn{1}{l}{}           & \multicolumn{1}{l}{}          & \multicolumn{1}{l}{}          \\
Ours                           & \cellcolor[HTML]{EA9999}26.96 & \cellcolor[HTML]{F9CB9C}0.8360 & \cellcolor[HTML]{EA9999}0.1890 & \cellcolor[HTML]{EA9999}85.68 & \cellcolor[HTML]{EA9999}63.82 & \cellcolor[HTML]{EA9999}25.13 & \cellcolor[HTML]{EA9999}0.8508 & \cellcolor[HTML]{EA9999}0.1567 & \cellcolor[HTML]{EA9999}96.42  & \cellcolor[HTML]{EA9999}54.29                      & \multicolumn{1}{l}{}          & \multicolumn{1}{l}{}           & \multicolumn{1}{l}{}           & \multicolumn{1}{l}{}          & \multicolumn{1}{l}{}          \\ \cline{1-11}
\end{tabular}
}
\label{tab:full_NVS}
\end{table*}

\end{document}